# Quadrotor Dead Reckoning with Multiple Inertial Sensors

D. Hurwitz and I. Klein

Autonomous Navigation and Sensor Fusion Lab
The Hatter Department of Marine Technologies,
University of Haifa,
Haifa, Israel

**Abstract**

Quadrotors are widely used for surveillance, mapping, and deliveries. In several scenarios the quadrotor operates in pure inertial navigation mode resulting in a navigation solution drift. To handle such situations and bind the navigation drift, the quadrotor dead reckoning (QDR) approach requires flying the quadrotor in a periodic trajectory. Then, using model or learning based approaches the quadrotor position vector can be estimated. We propose to use multiple inertial measurement units (MIMU) to improve the positioning accuracy of the QDR approach. Several methods to utilize MIMU data in a deep learning framework are derived and evaluated. Field experiments were conducted to validate the proposed approach and show its benefits.

## 1. Introduction

The use of quadrotors has significantly increased in the past decade in applications such as construction, delivery, transportation, surveillance, and more [1,2]. A quadrotor requires an accurate navigation system in order to perform challenging tasks. Global navigation satellite systems (GNSS) receivers combined with inertial navigation systems (INS) are commonly used for quadrotor navigation. INS/GNSS fusion provides accurate position and velocity information suitable for a wide range of applications [3,4]. When GNSS readings are unavailable (such as in urban canyons), pure inertial navigation is the only available navigation approach. Due to noise and errors in the inertial sensors, the navigation solution drifts over time [5].

Quadrotor dead reckoning (QDR) was recently proposed to handle pure inertial navigation situations [6]. In the same way that pedestrian dead reckoning (PDR) uses the wearable inertial sensors to determine the position of the pedestrian [7], quadrotor dead reckoning uses a predictable periodic motion trajectory instead of a straight line trajectory to calculate the location of the quadrotor. By doing this, the peak-to-peak change in distance of the quadrotor can be estimated, similar to how step-length is detected and estimated in PDR.

Motivated by recent applications of deep learning into navigation applications [8,9], we proposed QuadNet, a hybrid deep-learning framework for quadrotor dead reckoning that uses only inertial sensor measurements to estimate three-dimensional position [10]. Using regression neural networks, QuadNet calculates the quadrotor's change in distance and altitude, and determines its heading using model-based equations. Thus, for QuadNet to provide a three-dimensional position vector, only inertial sensor readings are required.

One of the means to improve pure inertial navigation is by utilizing other inertial sensors types such as angular accelerometers [11] or by designing different inertial accelerometers like gyro free configurations [12] or multiple inertial measurement units (MIMU) [13].

Recently, a comprehensive and thorough paper summarized the motivation for using MIMU [14]. Those include higher accuracy, reliability and dynamic measurement range than a single IMU. Additionally, MIMU architecture can estimate angular acceleration directly from accelerometer data as well as estimate angular motion from accelerometer data. In [15], an array of 192 sensors was used to derive closed form empirical solutions for coarse alignment, stationary calibration, and position accuracy of a MIMU system. Later, an approach for compensating the influence of position change in the center of gravity using MIMU was proposed [16]. MIMU were also used in pedestrian dead reckoning with shoe mounted sensors to improve the positioning of a walking pedestrian in an indoor environment [17, 18]. Recently, a visual-inertial navigation with multiple cameras and MIMU was proposed for mobile robots [19]. Fusion approaches for MIMU and external sensors are also a topic of interest. The most common approach is known as a virtual IMU, where all IMUs in the array a transformed to a single IMU [20,21]. Recently, the augmented virtual filter was proposed to allow the estimation of all inertial sensor error states in the MIMU system [22]. In the federated MIMU fusion approaches, each IMU in the array is treated separately, and then combined used a parameter estimation approach [23,24].

In this work, we propose to use MIMU with the QuadNet architecture to improve the quadrotor positioning accuracy. Several methods to utilize the MIMU data in a deep learning framework are proposed and examined. To validate our proposed approaches, field experiments were conducted using DJI's Phantom 4 quadrotor. This quadrotor has a RTK GNSS receiver providing our ground-truth trajectories. In addition, the quadrotor was equipped with four Xsens DOT IMUs to create the MIMU dataset for evaluations.

Our results shows that using four IMUs improves the position accuracy of a single IMU and also the benefits of MIMU data in the dataset. To promote further research in the field, this dataset was made publicly available under the ANSFL GitHub at: https://github.com/ansfl/Quadrotor-Dead-Reckoning-with-Multiple-Inertial-Sensors

The rest of the paper is organized as follows: Section 2 presents the model-based quadrotor dead reckoning approach. Section 3 describes the QuadNet framework and our proposed approach. Section 4 describes the experimental results while Section 5 gives the conclusions of this study.

## 2. Model-Based Quadrotor Dead Reckoning

The QDR approach was proposed as a means of dealing with quadrotor pure inertial navigation scenarios [6]. Essentially, the idea was to fly the quadrotor in a periodic motion trajectory (instead of a straight line), to mimic a walking pedestrian for the purpose of enabling PDR approaches to be conducted. Figure 1 illustrates a straight line trajectory used in situations of pure inertial navigation as well as a periodic QDR trajectory.

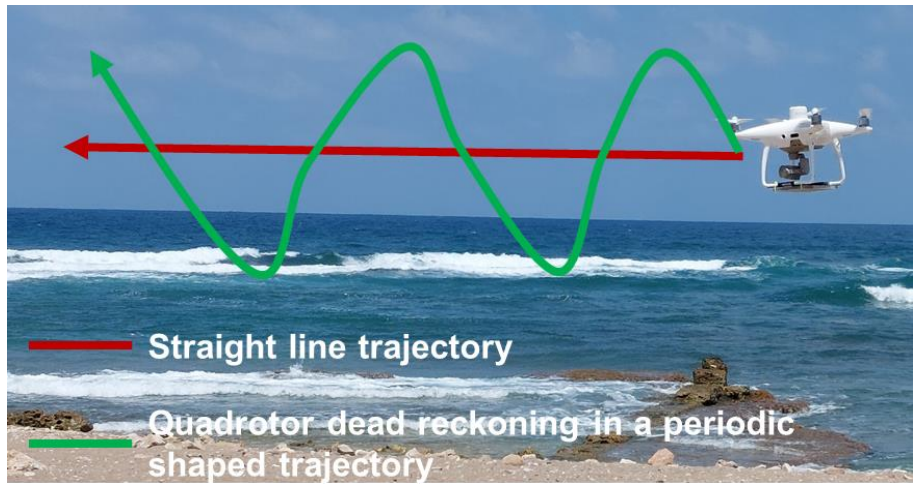

Figure 1. An illustration of a QDR periodic motion trajectory

Accelerometer readings are used to detect peaks during quadrotor motion, followed by Weinberg distance estimation. The orientation of the quadrotor is calculated in the same manner as in a traditional INS. Then, instead of using the inertial navigation equations, the QDR approach calculates the quadrotor's horizontal position based on the current p2p distance, heading, and initial position as shown in Figure 2.

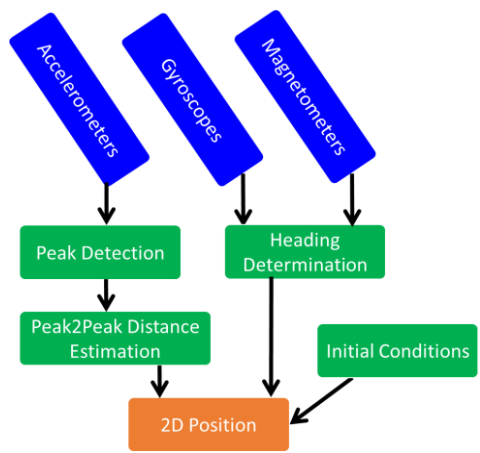

Figure 2. Block diagram of the QDR approach.

## 3. QuadNet with Multiple-Inertial Sensors

This section describes the QuadNet framework and its network architecture. Finally, our QuadNet MIMU approach is presented.

*3.1 QuadNet Framework*

Although QDR managed to improve pure inertial performance it can provide a 2D position solution only between two successive peaks, cannot estimate the quadrotor altitude, and is very sensitive to the model gain. To overcome some of the model-based QDR approach, QuadNet a hybrid learning approach was proposed [10]. QuadNet combines neural networks (NNs) and model-based equations. These networks are used to solve complex problems requiring a deep understanding of intricate relationships between a large number of interdependent variables or the discovery of hidden patterns in data. As in QDR, QuadNet requires the quadrotor to fly on a periodic trajectory. NN algorithms are utilized to estimate the quadrotor position vector. To that end, the inertial readings are plugged into a regression model in an end to end fashion. This is done to regress the change in distance and altitude of the quadrotor. In parallel, attitude and heading reference system algorithms are used to determine the heading angle. The change in distance, altitude, and heading are plugged into the equations of motion to produce the quadrotor 3D position, as presented in Figure 3.

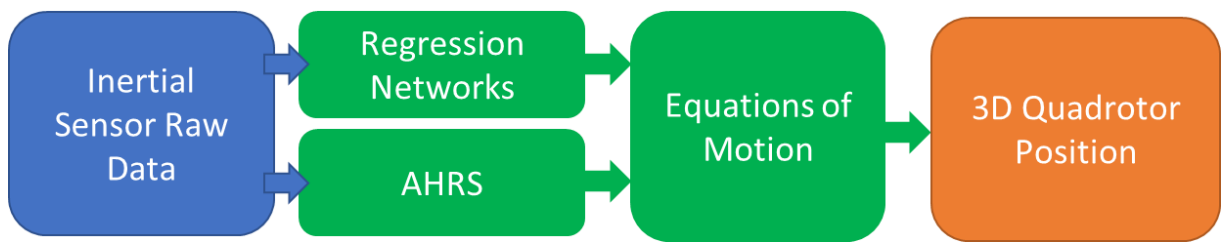

Figure 3. Block diagram of the QuadNet approach.

*3.2 Regression Network*

QuadNet is a deep neural network which consists of seven one-dimensional convolution neural networks (CNN) and three fully connected (FC) as illustrated in Figure 4. The input to the network is the raw inertial (accelerometer and gyroscope) readings and the output of the network is the change in distance or height. For quadcopter applications we found due their dynamics that a second of accelerometer and gyroscopes recordings window size holds most of the information, in our case it consists of 120 samples as indicated in the Figure.

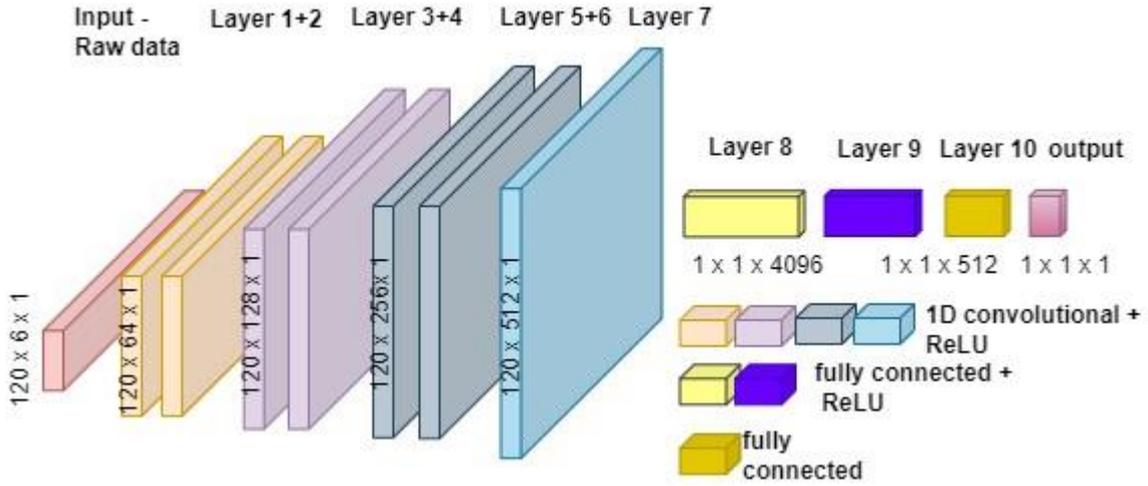

Figure 4 QuadNet architecture consists of 1D-CNN layers, used for feature extraction, and fully connected layers to output the change in distance or height.

The use of CNN and FC offers the advantages of both layers. CNN layers significantly reduces the number of parameters compared to fully-connected layers, allows to build larger networks, train them easily, and avoid overfitting. CNNs are also robustness to noise and are able to reason about the spatial relationships between different measurements. FC layers, on the other hand, are able to learn global features.

The convolution is defined by:

$$(x * W)(t) := \int_{-\infty}^{\infty} x(\tau) W(t - \tau) d\mathrm{T} \tag{1}$$

where $x$ is the input, $W$ is the filter matrix (weights), and $(*)$ is the convolution operator. The filter matrix is referred to as the weights matrix, which is updated during the training process. QuadNet has nine hidden layers between the input and output layers. Denoting $x$ as the input layer, the first hidden layer is defined by:

$$h^{(1)} = g^{(1)}\left(W^{(1)^T} x + b^{(1)}\right) \tag{2}$$

where $b$ is the bias vector, and $g$ is a nonlinear activation function. In following layers, the weights and biases create a mapping between the neurons in the current layer and the neurons from the previous layer, and the activation function allows the model to predict a range of cases that are not linear in nature. Here, the commonly used activation function, the rectified linear unit (ReLU) [25], is employed:

$$g(z) = max\{0, z\} \tag{3}$$

where *z* is the input to the activation function. The *i*-th layer is defined similar to (2), replacing the input by the output of the *i-1* layer:

$$h^{(i)} = g^{(i)}(W^{(i)^T} h^{(i-1)} + b^{(i)}) \tag{4}$$

*3.3 Loss Function*

The result of the output layer is compared to the ground-truth (GT) values by using a loss function. The GT values which were driven from a RTK-GNSS while the distance labels were a $l_2$ norm of the planner position. The minimization of the loss function is performed by the back-propagation process, where the loss function is derived and the weights in (2) and (4) are updated accordingly. Those two processes continue until a minimum is reached. For the problem at hand, the goal is to regress the change in distance or height.

To avoid large errors, we penalized more significantly large error by using quadratically penalization, enabling the network to estimate the desired output in a variety of conditions. To that end, the mean squared error (MSE) loss function is employed:

$$L(y_i, \hat{y}_i) = \frac{1}{N}\sum_{i=1}^{N}(y_i - \hat{y}_i)^2 \tag{5}$$

where N is the number of examples, $y_i$ is the GT value observed at time i, and $\hat{y}_i$ is the estimated value observed at time *i*.

*3.3 Multiple IMUs QuadNet*

We consider a MIMU array with *n* aligned IMUs. That is, all *x*-axis in the *n* IMUs are pointing to the same direction. The same applies for the *y* and *z* axes. As a results, the MIMU array has *n* accelerometers and *n* gyroscopes pointing to the same direction in each of the sensor frame coordinates making a total of *6n* inertial sensors.

The QuadNet network described in Section 3.2, holds several nonlinear operations. From a navigation perspective, as the inertial reading are proceed within this network, we consider two possibilities for the MIMU applications:

1) **Raw data average (RDA):** In this approach, the QuadNet network receives as input an average IMU readings obtained from the *n* aligned IMUs. To that end, an average operator is applied on each axis of the accelerometers and gyroscopes. The outcome of the network is the regression result. The RDA approach is illustrated in Figure 5.

2) **After regression average (ARA):** Here, each IMU readings is plugged to its QuadNet network, resulting in *n* networks, each outputting a regression estimate. Then, an average operator is applied on the regression estimate to give the final regression value.

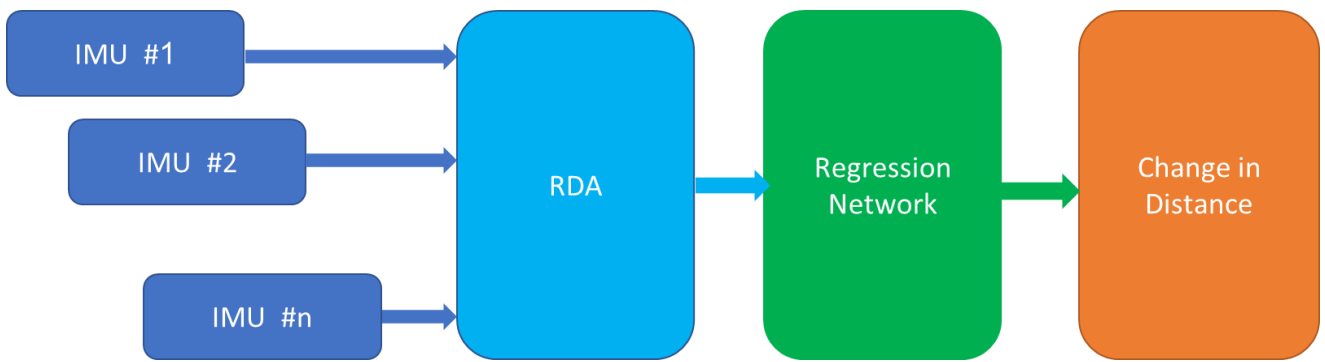

Figure 5. The RDA approach requires multiple IMUs and a single regression network.

Obviously, RDA approach requires less computational load as only a single network is required compared to n networks in the ARA approach. However, when only few IMUs are present in the MIMU array a trade-off between computational load and accuracy may raise. In addition from a practical point of view, a minimum training set is preferable as it will require less resources. Thus, we shall examine also the dataset size influence on the performance of both RDA and ARA approaches.

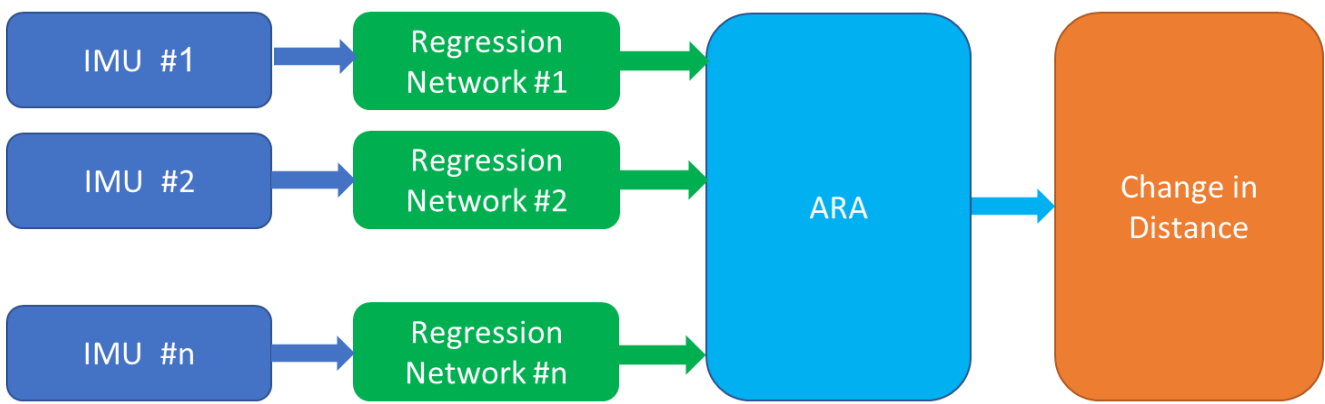

Figure 6. The ARA approach requires multiple IMUs and a multiple regression networks.

## 4. Analysis and Experimental Results

In this section we compare the baseline approach pure inertial navigation in straight line trajectories, to pure inertial navigation using QDR trajectories and MIMUs on an experimental dataset recording.

*4.1 Experimental Setup*

We employed a DJI phantom 4 RTK drone, and equipped it with four Xsens Dot sensors as shown in Figure 7. The GT data was taken from the quadrotor while Xsens Dot were the units under test. Xsens DOT sensor is small and lightweight sensor with an easy setup using BLE 5.0 1. The Xsens dot is based on a BOSCH BNO055 MEMS with 9DOF (Degrees Of Freedom) IMU - Measuring three-axes accelerometer three-axes gyroscope, and three-axes magnetometer. For our analysis only the accelerometer and gyroscope measurements were employed.

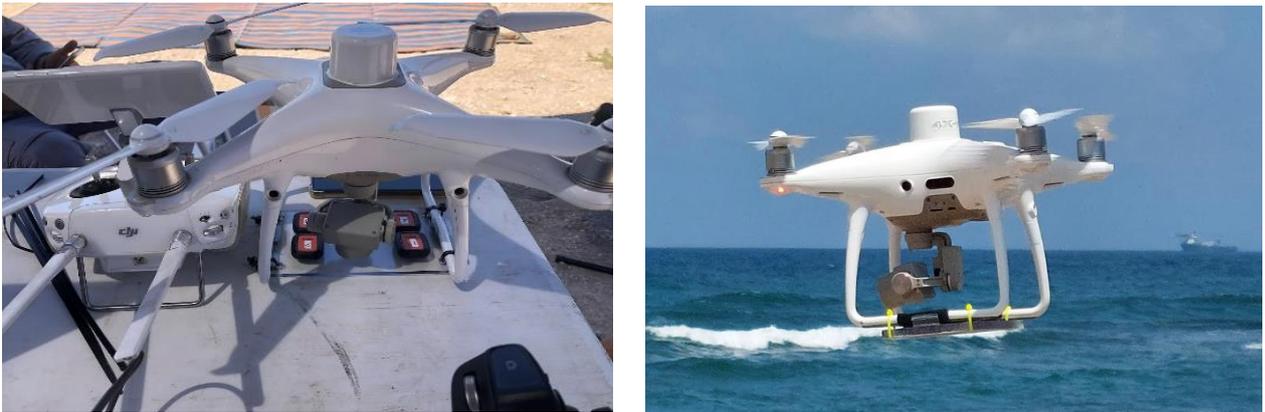

Figure 7. The quadrotor equipped with four IMUs in our field experiment setup.

*4.2 Dataset*

The datasets include the following recordings:

- **Straight lines trajectories**: four trajectories with a total time duration of 66 seconds while the quadrotor had an average speed of 5.4 m/s.

- **Horizontal periodic QDR trajectories**: 27 trajectories with a total time duration of 16.1 minutes while the quadrotor had an average speed of 3.7 m/s.

- **Vertical periodic QDR trajectories**: 31 trajectories with a total time duration of 12.6 minutes while the quadrotor had an average speed of 4.5 m/s.

In each trajectory includes 4 IMUs recording were made with a RTK-GNSS measurement severed as the ground truth (GT). In the vertical periodic QDR trajectories 17 out of the 31 trajectories included a single IMU due to IMU failure during the experiments. An example of vertical and horizontal periodic QDR trajectories is given in Figure 8.

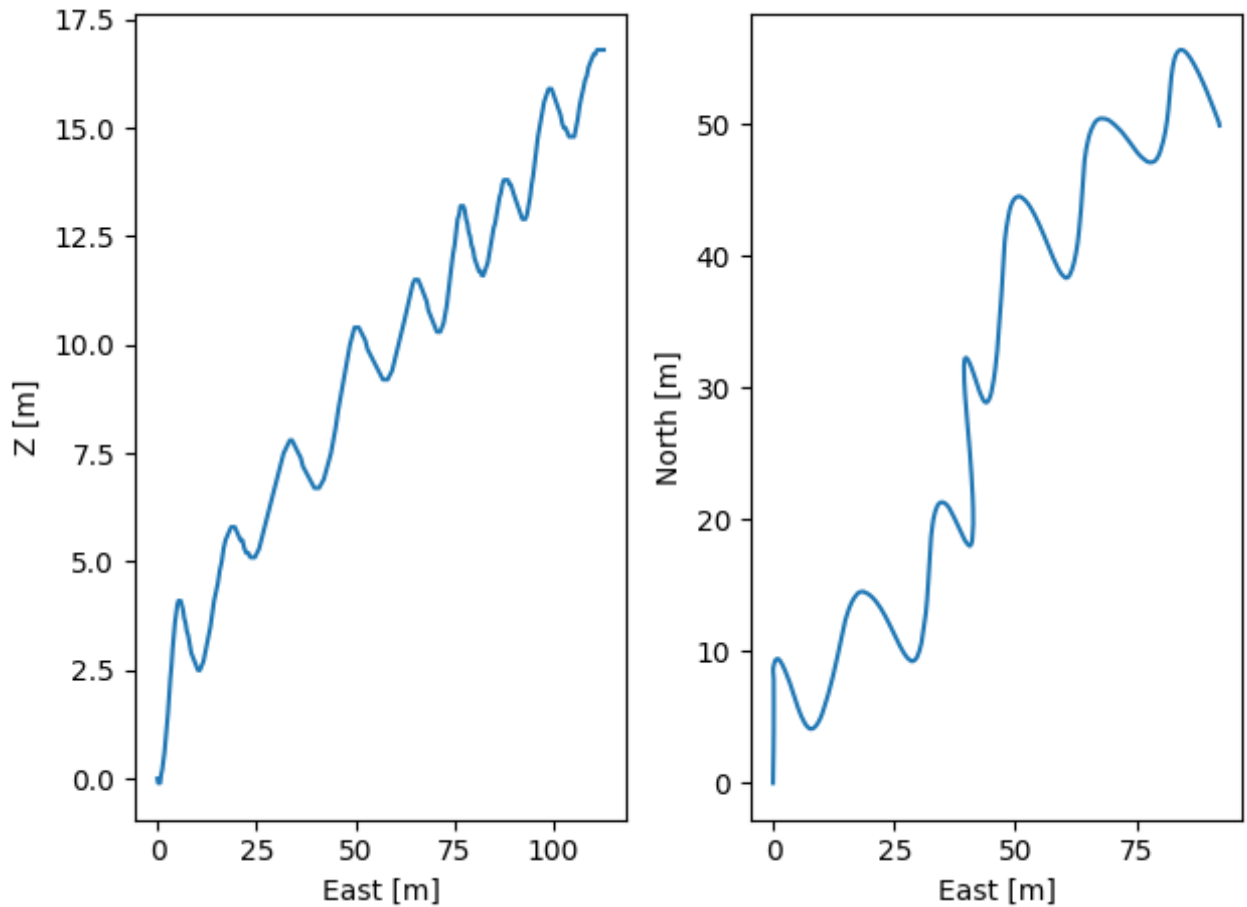

Figure 8. An example of QDR trajectories (a) vertical (side view) and (b) horizontal (top view).

The straight line trajectories were used to evaluate the pure inertial solution using the model-based INS equations (baseline approach). To train our QuadNet network a single arbitrary trajectory was set as the test dataset with the following division made:

- **D1: Horizontal Dataset**. Train: all trajectories of IMU #1 except trajectory 4. Test: IMU #1, trajectory 4

- **D2: Vertical Dataset**. Train : all trajectories of IMU #1 except trajectory 9 Test: trajectory 9

- **D3: Horizontal Dataset**. Train: all trajectories of all IMUs except trajectory 4. Test: trajectory 4

- **D4: Vertical Dataset**. Train : all trajectories of all IMUs except trajectory 9. Test: trajectory 9

*4.3 Experimental Results*

*4.3.1. Performance metrics:*

The purposed method is evaluated by three metrices, the root mean square error (RMSE) metric, the max error, and the standard deviation of the entire trajectory estimation. The RMSE is defined by:

$$\mathbf{RMSE}(x_i, \hat{x}_i) = \sqrt{\frac{\Sigma_{i=1}^{N}(x_i - \hat{x}_i)^2}{N}} \tag{1}$$

where *N* is the number of samples, $x_i$ is the GT distance/height observed at time *i*, and $\hat{x}_i$ is the estimated distance/height observed at time *i*.

*4.3.2. Straight Line - INS (baseline)*

In Table 1 the INS RMSE of a single IMU (out of the four IMUs) in the four straight lines trajectories. As expected, the errors are in the scale of tens of meters almost equals to the trajectory length. This results for many applications especially for quadcopters can be devastating and may cause the quadcopter to crash.

Table 1: Performance for straight lines trajectories using a single IMU.

| Scenario | RMSE [m] | Max Error [m] | STD [m] | Trajectory length [m] |
|---|---|---|---|---|
| Straight line #1 | 384 | 938 | 272 | 67 |
| Straight line #2 | 221 | 519 | 155 | 92 |
| Straight line #3 | 210 | 490 | 146 | 96 |
| Straight line #4 | 88 | 220 | 64 | 101 |
| Mean | 226 | 542 | 159 | 89 |

The results for the MIMU is presented in Table 2. We followed the RDA approach and averaged the IMUs readings before calculating the trajectory.

Table 2: Performance for straight lines trajectories using four IMUs with the RDA approach.

| Scenario | RMSE [m] | Max Error [m] | STD [m] | Trajectory length [m] |
|---|---|---|---|---|
| Straight line #1 | 266 | 625 | 183 | 67 |
| Straight line #2 | 195 | 452 | 136 | 92 |
| Straight line #3 | 179 | 419 | 125 | 96 |
| Straight line #4 | 76 | 175 | 53 | 101 |
| Mean | 179 | 418 | 124 | 89 |

Results show an advantage of using multiple IMUs as the mean RMSE of the four trajectories was reduce by 20%. Yet, the RMSE value presents a large error of 179 meters, not acceptable for quadrotor applications.

### 4.3.3. Horizontal QDR

The change in distance estimation performance as a function of number of IMUs in the MIMU configuration is presented in Table 3 for the D1 Horizontal Dataset test trajectory (a single trajectory with length of 117 meters) and the RDA approach. There, each row represent average of IMUs as an input to the QuadNet (RDA) and an average on the results of all available IMU combinations. For example, line two with Two IMUs shows the average of the following six IMU combinations: {1,2}, {1,3}, {1,4}, {2,3}, {2,4}, and {3,4}.

It is clearly seen in Table 3 that QuadNet performance greatly improved the pure inertial baseline solution. In addition, as the number of IMUs is increased the performance is improved. For example, an improvement of 13% when using four IMUs instead of a single IMU.

Table 3: Performance as a function of number of IMUs in QuadNet distance estimation trained on D1 Horizontal Dataset while using the RDA approach for testing.

| Number of IMUs | RMSE [m] | Max error [m] | STD [m] |
|---|---|---|---|
| One | 0.78 | 1.64 | 0.42 |
| Two | 0.72 | 1.49 | 0.41 |
| Three | 0.70 | 1.44 | 0.40 |
| Four | 0.68 | 1.48 | 0.40 |

We follow the same steps only for the ARA approach. Notice, that as we use all possible combinations and present their average, the ARA approach is equivalent to the results of single IMU shown in Table 2 for RDA approach. Thus, using RDA with four IMUs shows better results and also requires less computational effort.

An example of the change in distance in each epoch (1 second) is present in Figure 7 showing the GT distance compared of the QuadNet regressed distance.

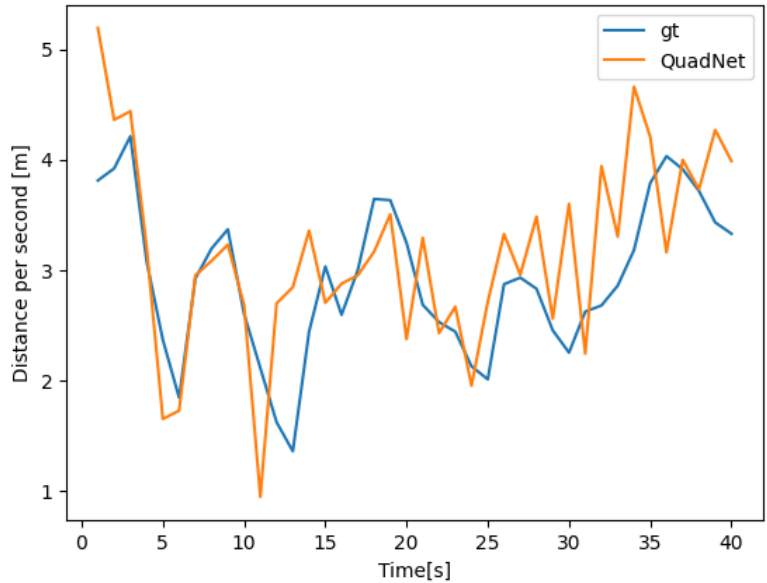

Figure 7 Change in distance in each epoch (1 second) showing GT and QuadNet distance values.

Next, the influence of the training dataset is evaluated as we repeat the analysis for the RDA approach on D3 Horizontal Dataset test trajectory. Table 4 presents the results showing almost the same performance regardless of the number of IMUs. Also, it shows an improvement of 28% compared to a single IMU and 15% compared to using four IMUs in the D1 dataset. Thus, when using MIMU data in the training set may reduce the need of using MIMU in the actual missions.

Table 4: Performance as a function of number of IMUs in QuadNet distance estimation trained on D3 Horizontal Dataset while using the RDA approach for testing.

| Number of IMUs | RMSE [m] | Max error [m] | STD [m] |
|---|---|---|---|
| One | 0.56 | 1.40 | 0.33 |
| Two | 0.58 | 1.40 | 0.34 |
| Three | 0.60 | 1.52 | 0.37 |
| Four | 0.58 | 1.64 | 0.38 |

### 4.3.4. Vertical QDR

The same procedure as in Section 4.3.3 is repeated here for the D2 vertical dataset with a test trajectory length of 99.4 meters. The trend of more IMUs improving distance estimation using the RDA approach is also evident in Table 5. An improvement of 17% was obtained when using four IMUs instead of a single IMU, which equals to the improvement of RDA over ARA.

Table 5: Performance as a function of number of IMUs in QuadNet distance estimation trained on D2 Vertical Dataset while using the RDA approach for testing.

| Number of IMUs | RMSE [m] | Max error [m] | STD [m] |
|---|---|---|---|
| Single | 1.70 | 3.70 | 0.91 |
| Two | 1.54 | 3.69 | 0.84 |
| Three | 1.46 | 3.61 | 0.82 |
| Four | 1.41 | 3.52 | 0.8 |

Next, the influence of the training dataset is evaluated as we repeat the analysis for the RDA approach on D4 Vertical Dataset test trajectory. As shown in Table 6, the performance has improved compared to training on dataset D2, as a result of the larger dataset D4. In contrast with the behavior in the horizontal case, Table 6 shows that as more IMUs are present in the test the performance improves, particularity by 15% when using four IMUs compared to a single IMU.

Table 6: Performance as a function of number of IMUs in QuadNet distance estimation trained on D4 Vertical Dataset while using the RDA approach for testing.

| Number of IMUs | RMSE [m] | Max error [m] | STD [m] |
|---|---|---|---|
| One | 1.34 | 2.95 | 0.81 |
| Two | 1.23 | 2.80 | 0.73 |
| Three | 1.17 | 2.80 | 0.71 |
| Four | 1.14 | 2.70 | 0.69 |

After examining the distance estimation performance, we focus on the change of altitude of the quadrotor. Using dataset D2, we examine the influence of the number of IMUs on the altitude regression accuracy. Table 7, shows that using more than a single IMU improve the performance. Yet, the major improvement of 11% was obtained when moving from a single IMU to two IMUs.

Table 7: Performance as a function of number of IMUs in QuadNet altitude estimation trained on D2 Vertical Dataset while using the RDA approach for testing.

| Number of IMUs | RMSE [m] | Max error [m] | STD [m] |
|---|---|---|---|
| One | 0.73 | 1.62 | 0.45 |
| Two | 0.65 | 1.44 | 0.38 |
| Three | 0.64 | 1.41 | 0.36 |
| Four | 0.63 | 1.37 | 0.35 |

The same characteristics occur when training on the D4 dataset. The results in Table 8 shows that the addition of more IMUs improve the performance compared to a single IMU, and again the major improvement is in the transition from a single IMU to two IMUs. Also, using this training set improved the RMSE of four IMUs by 21%.

Table 8: Performance as a function of number of IMUs in QuadNet altitude estimation trained on D4 Vertical Dataset while using the RDA approach for testing.

| Number of IMUs | RMSE [m] | Max error [m] | STD [m] |
|---|---|---|---|
| One | 0.59 | 1.60 | 0.41 |
| Two | 0.52 | 1.36 | 0.35 |
| Three | 0.51 | 1.27 | 0.35 |
| Four | 0.50 | 1.25 | 0.34 |

**5. Conclusion**

Using MIMU with the QuadNet approach was suggested to improve the quadrotor positioning accuracy. To that end, a unique dataset was recorded using DJI's Phantom 4 RTK quadrotor which was equipped with four Xsens DOT IMUs. To promote further research in the field, this dataset was made publicly available under the ANSFL GitHub at: https://github.com/ansfl/Quadrotor-Dead-Reckoning-with-Multiple-Inertial-Sensors

Two different approaches for MIMU utilization were examined, namely the RDA and ARA. Our results show that the RDA approach obtained the better performance compared to ARA. Also, RDA requires only a single network while ARA requires a network for each IMU. Thus, in terms of accuracy and computational load RDA is preferred.

The RDA approach applied for horizontal or vertical QDR trajectories greatly improved the baseline pure inertial navigation. In addition, for both horizontal and vertical QDR trajectories trained on single IMU dataset, as the number of IMUs used in the testing, the performance was improved. Specifically for when using four IMUs instead of a single IMU the improvement was 13-17%.

When training on MIMU data the accuracy of the estimation improved regardless to the number of IMUs used in the testing. For the horizontal scenarios, the results showed almost the same performance regardless of the number of IMUs. On the other hand, for the vertical scenarios, both in distance and altitude estimation, the accuracy improved as the number of IMUs was increased,

To summarize, we showed that using MIMU improves the performance of the QuadNet approach. As the number of IMUs increases also does the performance. In addition, using a training dataset composing of MIMU date contributes to improving the accuracy compared to a single IMU data set.

**Acknowledgement**

This research was partially supported by the Israeli Ministry of Science, Technology, and Space under grant number 1001575651.

**References**

[1] M. Idrissi, M. Salami, and F. Annaz, "A review of quadrotor unmanned aerial vehicles: applications, architectural design and control algorithms", Journal of Intelligent and Robotic Systems, vol. 104(2), pp. 22, 2022.

[2] G. Sonugur, "A Review of quadrotor UAV: Control and SLAM methodologies ranging from conventional to innovative approaches", Robotics and Autonomous Systems, p.104342, 2022.

[3] G. Zhang, L. T. and Hsu, "Intelligent GNSS/INS integrated navigation system for a commercial UAV flight control system", Aerospace science and technology, vol. 80, pp. 368-380, 2018

[4] A. Borko, A., I. Klein, and G. Even-Tzur, "GNSS/INS fusion with virtual lever-arm measurements", Sensors, vol. 18(7), pp. 2228, 2018.

[5] D. Titterton and J. L. Weston, "Strapdown Inertial Navigation Technology", Reston, VA, USA: American Institute of Aeronautics and Astronautics and the Institution of Electrical Engineers, 2004.

[6] A. Shurin and I. Klein, "QDR: A Quadrotor Dead Reckoning Framework", IEEE Access, vol. 8, pp. 204433–204440, 2020.

[7] X. Hou and J. Bergmann, "Pedestrian Dead Reckoning With Wearable Sensors: A Systematic Review", IEEE Sensors Journal, vol. 21(1), pp. 143-152, 2021.

[8] N. Cohen, and I. Klein, "Inertial Navigation Meets Deep Learning: A Survey of Current Trends and Future Directions", arXiv preprint arXiv:2307.00014, 2023.

[9] I. Klein, "Data-driven meets navigation: Concepts, models, and experimental validation", DGON Inertial Sensors and Systems (ISS), pp. 1-21, 2022.

[10] A. Shurin and I. Klein, "QuadNet: A hybrid framework for quadrotor dead reckoning", Sensors, vol. 22(4), pp. 1426, 2022.

[11] U. Nusbaum, I. Rusnak, and I. Klein, "Angular accelerometer-based inertial navigation system", Navigation, vol. 66(4), pp. 681-693, 2019.

[12] M. Maynard, and V. Vikas, "Angular Velocity Estimation Using Non-Coplanar Accelerometer Array", IEEE Sensors Journal, vol. 21(20), pp. 23452-23459, 2021.

[13] J. B. Bancroft, and G. Lachapelle, "Data fusion algorithms for multiple inertial measurement units", Sensors, vol. 11(7), pp. 6771-6798, 2011.